\pdfoutput=1 
\documentclass[10pt,twocolumn,letterpaper]{article}

\usepackage[pagenumbers]{cvpr} 

\usepackage[dvipsnames]{xcolor}

\usepackage{amssymb}
\usepackage{pifont}

\usepackage{multirow}
\usepackage{adjustbox}
\usepackage{pgfplots}
\pgfplotsset{compat=newest}
\usepackage{colortbl}
\usepackage[accsupp]{axessibility}
\usepackage{siunitx}

\definecolor{cvprblue}{rgb}{0.21,0.49,0.74}
\usepackage[pagebackref,breaklinks,colorlinks,citecolor=cvprblue,bookmarks=false]{hyperref}

\def\paperID{5}
\def\confName{CVPR}
\def\confYear{2024}

\title{How to Benchmark Vision Foundation Models for Semantic Segmentation?}

\author{Tommie Kerssies, Daan de Geus, Gijs Dubbelman\\
Eindhoven University of Technology \\
{\tt\small \{t.kerssies, d.c.d.geus, g.dubbelman\}@tue.nl}
}

\begin{document}
\maketitle
\begin{abstract}
    Recent vision foundation models (VFMs) have demonstrated proficiency in various
tasks but require supervised fine-tuning to perform the task of semantic
segmentation effectively. Benchmarking their performance is essential for
selecting current models and guiding future model developments for this task.
The lack of a standardized benchmark complicates comparisons. Therefore, the
primary objective of this paper is to study how VFMs should be benchmarked for
semantic segmentation. To do so, various VFMs are fine-tuned under various
settings, and the impact of individual settings on the performance ranking and
training time is assessed. Based on the results, the recommendation is to
fine-tune the ViT-B variants of VFMs with a $16\times16$ patch size and a
linear decoder, as these settings are representative of using a larger model,
more advanced decoder and smaller patch size, while reducing training time by
more than 13 times. Using multiple datasets for training and evaluation is also
recommended, as the performance ranking across datasets and domain shifts
varies. Linear probing, a common practice for some VFMs, is not recommended, as
it is not representative of end-to-end fine-tuning. The benchmarking setup
recommended in this paper enables a performance analysis of VFMs for semantic
segmentation. The findings of such an analysis reveal that pretraining with
promptable segmentation is not beneficial, whereas masked image modeling (MIM)
with abstract representations is crucial, even more important than the type of
supervision used. The code for efficiently fine-tuning VFMs for semantic
segmentation can be accessed through the project
page~\footnote{\url{https://tue-mps.github.io/benchmark-vfm-ss/}}.
\end{abstract}

\section{Introduction}
\pgfplotstableread{
      Choice                             Correlation
            {Mask2Former decoder}                   0.87
            {ViT-L}                         0.87
            {$8\times8$ patch size}                  0.78
            {PASCAL VOC}                    0.78
            {GTA V$\rightarrow$Cityscapes}  0.64
            {Cityscapes}                    0.56
            {Linear probing}                0.47
}\modeltable

\begin{figure}[t]
      \footnotesize
      \centering
      \scalebox{0.85}{
            \begin{tikzpicture}
                  \begin{axis}[
                              ybar,
                              enlargelimits=0.15,
                              symbolic x coords={Mask2Former decoder, ViT-L, $8\times8$ patch size, PASCAL VOC, {GTA V$\rightarrow$Cityscapes}, Cityscapes, Linear probing},
                              xtick=data,
                              xticklabel style={rotate=35, anchor=east},
                              nodes near coords,
                              every node near coord/.append style={yshift=5pt},
                              ylabel={Kendall rank correlation coefficient ($\tau$)},
                              grid=major
                        ]
                        \addplot+[error bars/.cd, y dir=both, y explicit] table [x=Choice, y=Correlation] {\modeltable};
                  \end{axis}
            \end{tikzpicture}
      }
      \caption{\textbf{Performance ranking impact of settings.}
            Kendall's $\tau$ is used to assess ranking similarity between VFMs under
            default settings (linear decoder, ViT-B, $16\times16$ patch size, ADE20K,
            end-to-end fine-tuning) and after changing individual settings, ranging from -1 for a
            reverse ranking to 1 for an identical ranking.}\label{fig:effect}
\end{figure}
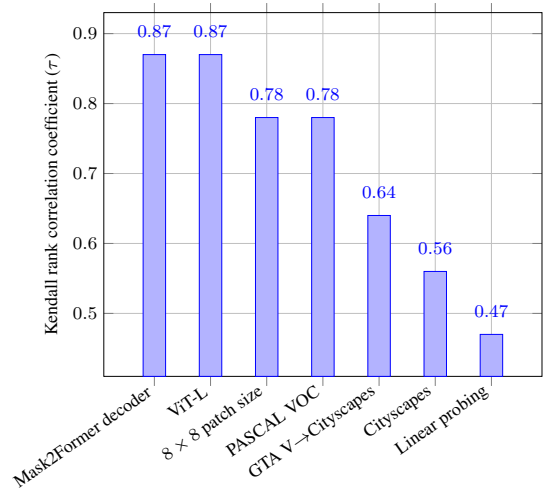

Semantic segmentation is the task of assigning a semantic class to each pixel
in an image. Training a model to perform this task requires a large amount of
images with semantic mask labels, presenting a significant challenge due to the
intensive labor associated with such
annotation~\cite{everingham2010pascal,cordts2016cityscapes}. Vision Foundation
Models (VFMs), \ie, vision models pretrained on broad datasets, offer a
solution to the labeling burden. A VFM acquires a fundamental understanding of
visual features from the pretraining task, such that it can be used for a large
variety of downstream tasks~\cite{bommasani2021opportunities}. By transfer
learning from the pretraining task to the downstream task of semantic
segmentation, the need for extensive annotation is
reduced~\cite{vemulapalli2023label,bensaid2024novel}. To gain insight into
which VFMs are most effective for the task of semantic segmentation, a
standardized benchmarking setup is crucial, but currently lacking. Therefore,
the primary objective of this paper is to study how VFMs should be benchmarked
for this task.

The Vision Transformer (ViT)~\cite{dosovitskiy2020image} is a simple
architecture which enables the use of global image context throughout the
network. Recent VFMs have mostly adopted the same ViT architecture with various
pretraining strategies~\cite{fang2023eva2, sun2023eva,oquab2023dinov2,
      wang2022image,zhai2023sigmoid, fang2023data, touvron2022deit, he2022masked,
      kirillov2023segment}. These VFMs demonstrate that better pretraining on more
data results in better downstream performance for various tasks. However, it is
unclear which pretraining strategies are most effective for the task of
semantic segmentation specifically. Although some VFMs were evaluated for this
task through supervised
fine-tuning~\cite{oquab2023dinov2,he2022masked,fang2023eva2,wang2022image,touvron2022deit},
a representative comparison is hindered by the lack of uniform evaluation
settings. The inconsistency stems from diverse experimental conditions,
including different downstream datasets, parameter freezing practices, number
of tokens per image, model dimensions, and decoders. The precise impact of
these factors on the performance ranking is not well understood.

This paper examines how variations in evaluation settings impact the
performance ranking of VFMs for semantic segmentation and what recommendations
can be made for a benchmarking setup that is most efficient while still being
representative, \ie, a setup that optimizes training efficiency while
accurately reflecting the performance ranking that would emerge under more
resource-intensive tuning. By improving efficiency, benchmarking new models on
new datasets is made accessible to a wider audience. Moreover, a representative
benchmarking setup ensures that performance comparisons are conducted on an
equitable basis, facilitating clear, unbiased insights into the efficacy of
various pretraining strategies for semantic segmentation.

In this paper, an efficient default benchmarking setup is used to fine-tune
various VFMs for semantic segmentation, establishing a baseline performance
ranking. Various individual benchmark settings are then changed to observe
their impact on the performance ranking, with an overview of the results
provided in Figure~\ref{fig:effect}. More details, and the recommendations that
can be made for benchmarking, are provided in Section~\ref{sec:impact}.
Finally, the recommended benchmarking setup enables a performance analysis of
VFMs for semantic segmentation, detailed in Section~\ref{sec:analysis}.

The primary contributions of this paper include:
\begin{itemize}
      \item An impact analysis of benchmark settings on the performance ranking of VFMs for
            semantic segmentation.
      \item A recommended setup for benchmarking VFMs for this task, which balances
            efficiency with representativeness.
      \item A performance analysis of VFMs for this task, using the recommended
            benchmarking setup.
      \item The code for efficiently fine-tuning VFMs for this task.
\end{itemize}
\section{Related work}

\quad\textbf{(V)FMs.} A foundation model (FM) is defined as a model that is pretrained on a broad
dataset, such that it can be used for a large variety of downstream
tasks~\cite{bommasani2021opportunities}. The concept of FMs was popularized in
the realm of natural language processing, with models like
BERT~\cite{devlin2018bert} and GPT~\cite{radford2018improving} demonstrating
remarkable capabilities in understanding and generating human language. In
computer vision, ImageNet~\cite{deng2009imagenet} pretraining stands as an
earlier example of a VFM, showcasing the power of transfer learning by
fine-tuning a broadly pretrained model for diverse applications.

Following the success of ImageNet pretraining, VFMs have evolved to embrace a
variety of pretraining strategies. Pretraining strategies for VFMs can be
broadly categorized into
fully-supervised~\cite{touvron2022deit,kirillov2023segment},
weakly-supervised~\cite{radford2021learning, gadre2023datacomp, fang2023data,
    xu2023demystifying,zhai2023sigmoid,cherti2023reproducible}, and
self-supervised~\cite{oquab2023dinov2, he2022masked, assran2023self} learning.
Some VFMs are specifically designed for supervised fine-tuning on downstream
tasks~\cite{fang2023eva2,oquab2023dinov2,wang2022image,
    touvron2022deit,he2022masked}, while others are designed for zero-shot
capabilities~\cite{radford2021learning, gadre2023datacomp, fang2023data,
    xu2023demystifying,zhai2023sigmoid,cherti2023reproducible}. Given the inferior
performance of zero-shot semantic segmentation methods compared to supervised
fine-tuning~\cite{wysoczanska2023clip, naeem2023silc}, current VFMs require
supervised fine-tuning to perform the task of semantic segmentation
effectively. Yet, it remains unclear which pretraining strategies are most
effective for this task.

\textbf{VFMs and semantic segmentation.} A representative comparison of the fine-tuned performance of VFMs for this task
is hindered by the lack of uniform evaluation settings. An overview of the
reported performance in the original papers of a selection of VFMs
for IN1K~\cite{deng2009imagenet} classification and ADE20K~\cite{zhou2017scene}
semantic segmentation is provided in Table~\ref{tab:reported}. Several data
points are missing, as they are not reported. Furthermore, the heterogeneity in
evaluation protocols complicates the comparison. It also remains ambiguous whether
IN1K classification performance is indicative of ADE20K semantic segmentation
performance.

\begin{table*}\footnotesize
    \centering
    \scalebox{0.85}{
        \begin{tabular}{lcccccc}
            \toprule
            \multicolumn{1}{c}{}                                     & \multicolumn{3}{c}{IN1K classification validation accuracy (\%)} &                         & \multicolumn{2}{c}{ADE20K semantic segmentation validation mIoU (\%)}                                               \\
            \cmidrule{2-4} \cmidrule{6-7}
                                                                     & Zero-shot                                                        & Linear probing          & End-to-end fine-tuning                                                &  & Linear probing & End-to-end fine-tuning* \\
            \midrule
            EVA-02~\cite{fang2023eva2}                               & --                                                               & --                      & 87.0 (196), 88.3 (1024)                                               &  & --             & 55.3 (1024)             \\
            EVA-02-CLIP~\cite{sun2023eva}                            & 74.7 (196)                                                       & --                      & --                                                                    &  & --             & --                      \\
            DINOv2~\cite{oquab2023dinov2}                            & --                                                               & 84.5 (256), 86.7 (1024) & 88.5 (256), 88.9 (1024)                                               &  & 47.3 (1369)    & --                      \\
            BEiT-3~\cite{wang2022image}                              & --                                                               & --                      & 85.4 (196)                                                            &  & --             & --                      \\
            SigLIP~\cite{zhai2023sigmoid}                            & 76.2 (196), 79.1 (1024)                                          & --                      & --                                                                    &  & --             & --                      \\
            DFN~\cite{fang2023data}                                  & 76.2 (196)                                                       & --                      & --                                                                    &  & --             & --                      \\
            DeiT III (IN21K$\rightarrow$IN1K)~\cite{touvron2022deit} & --                                                               & --                      & 86.7 (576)                                                            &  & --             & --                      \\
            DeiT III (IN1K)~\cite{touvron2022deit}                   & --                                                               & --                      & 85.0 (576)                                                            &  & --             & --                      \\
            MAE~\cite{he2022masked}                                  & --                                                               & 68.0 (196)              & 83.6 (196)                                                            &  & --             & 48.1 (1024)             \\
            SAM~\cite{kirillov2023segment}                           & --                                                               & --                      & --                                                                    &  & --             & --                      \\
            \bottomrule
        \end{tabular}
    }
    \caption{\textbf{Reported performance of VFMs.}
        Reported performance in the original papers of a selection of VFMs for IN1K classification
        and ADE20K semantic segmentation for the ViT-B variants with
        highest pretraining image size. Parentheses show number of tokens per image
        for evaluation. -- denotes the performance is not reported, * denotes using
        the UPerNet decoder~\cite{xiao2018unified}.}\label{tab:reported}
\end{table*}

There is one VFM in Table~\ref{tab:reported} specifically designed for
segmentation. SAM~\cite{kirillov2023segment} solves the task of promptable
segmentation and introduces the first large-scale segmentation dataset (SA-1B).
However, the masks in this dataset lack semantic information and are not
semantically consistent, \ie, they arbitrarily belong to, \eg, whole objects,
parts or subparts. Consequently, SAM does not support semantic prompts, and
requires supervised fine-tuning to perform semantic
segmentation~\cite{wang2023sam, ranzinger2023radio}. It is unclear whether SAM
would perform better for semantic segmentation than VFMs that were not
specifically designed for segmentation.

\textbf{VFM benchmarking for semantic segmentation.} A recent paper~\cite{bensaid2024novel} investigates the adaptability of various
VFMs for few-shot semantic segmentation through linear probing as well as
end-to-end fine-tuning. However, the paper is limited to the 1-shot setting,
while settings with more downstream data are relevant in many applications. It
is unclear how the results for the 1-shot setting would generalize to more
downstream data. In another recent paper, AM-RADIO~\cite{ranzinger2023radio}
combines multiple VFMs into a single model, and includes a benchmarking process
that covers multiple downstream tasks including semantic segmentation. However,
the evaluation is limited to linear probing, while end-to-end fine-tuning may
yield better performance. It is unclear how representative the results for
linear probing are for end-to-end fine-tuning, warranting further
investigation.

In summary, it is currently not possible to identify the best VFM for semantic
segmentation. This underscores the critical need for a well-designed,
standardized benchmarking setup for the assessment of the performance of VFMs
for this task. To address this need, this paper recommends a benchmarking setup
that facilitates the comparison between existing VFMs, and guides the
development of future VFMs towards enhanced performance for this task.
\section{Benchmarking setup}
\subsection{Models}
\label{sec:models}

To study how VFMs should be benchmarked for semantic segmentation, a diverse
set of VFMs is selected for their representation of the latest advancements in
the field, with varying training data sources, objectives, supervision methods,
and pretraining image sizes, as shown in Table~\ref{tab:models}. The models
share the same fundamental ViT~\cite{dosovitskiy2020image} architecture and the
variants with the highest pretraining image size are selected for benchmarking.

\begin{table*}\footnotesize
    \centering
    \scalebox{0.85}{
        \begin{tabular}{lllll}
            \toprule
            Name                                                     & Data                                                                                                            & Objective                               & Supervision                         & Tokens               \\
            \midrule
            EVA-02~\cite{fang2023eva2}                               & IN-21K~\cite{deng2009imagenet} (B) / Merged-38M~\cite{fang2023eva2} (L)                                         & MIM                                     & CLIP teacher                        & 196                  \\
            EVA-02-CLIP~\cite{sun2023eva}                            & IN-21K~\cite{deng2009imagenet} (B) / Merged-38M~\cite{fang2023eva2} (L)$\rightarrow$Merged-2B~\cite{sun2023eva} & MIM$\rightarrow$CLIP                    & CLIP teacher$\rightarrow$texts      & 196$\rightarrow$196  \\
            DINOv2~\cite{oquab2023dinov2}                            & LVD-142M~\cite{oquab2023dinov2}                                                                                 & MIM, discrimination                     & --                                  & 256$\rightarrow$1369 \\
            BEiT-3~\cite{wang2022image}                              & IN-21K~\cite{deng2009imagenet}, image-text~\cite{wang2022image}, text~\cite{wang2022image}                      & MIM, MLM                                & CLIP teacher, texts                 & 196                  \\
            SigLIP~\cite{zhai2023sigmoid}                            & WebLI~\cite{iscen2023improving}                                                                                 & LIP                                     & Texts                               & 1024 (B) / 576 (L)   \\
            DFN~\cite{fang2023data}                                  & DFN-2B~\cite{fang2023data}                                                                                      & CLIP                                    & Texts                               & 196                  \\
            DeiT III (IN21K$\rightarrow$IN1K)~\cite{touvron2022deit} & IN-21K~\cite{deng2009imagenet}$\rightarrow$IN1K~\cite{deng2009imagenet}                                         & Classification                          & Classes                             & 576                  \\
            DeiT III (IN1K)~\cite{touvron2022deit}                   & IN1K~\cite{deng2009imagenet}                                                                                    & Classification                          & Classes                             & 576                  \\
            MAE~\cite{he2022masked}                                  & IN1K~\cite{deng2009imagenet}                                                                                    & MAE                                     & --                                  & 196                  \\
            SAM~\cite{kirillov2023segment}                           & IN1K~\cite{deng2009imagenet}$\rightarrow$SA-1B~\cite{kirillov2023segment}                                       & MAE$\rightarrow$promptable segmentation & --$\rightarrow$class-agnostic masks & 196$\rightarrow$4096 \\
            \bottomrule
        \end{tabular}
    }
    \caption{\textbf{Overview of VFMs.} Comparison by pretraining data source, learning objective, supervision type, and number of tokens per image.
        $\rightarrow$: transfer learning.
    }\label{tab:models}
\end{table*}

\subsection{Settings}
\label{sec:settings}

\quad\textbf{Freezing the encoder.} By freezing the encoder and using a linear layer as decoder, commonly
referred to as linear probing, the encoder is used as a fixed feature
extractor, while only the parameters of the linear layer are learned. In
contrast, end-to-end fine-tuning allows both the encoder and decoder to be
updated. Linear probing would be ideal for adapting a VFM to the task
of semantic segmentation, as a VFM should have learned a rich set of features
in pretraining that need no fine-tuning. As such, some VFMs
solely evaluate semantic segmentation performance through linear
probing~\cite{oquab2023dinov2, assran2023self}. To assess whether freezing the
encoder impacts the performance ranking, the analysis compares linear probing
to end-to-end fine-tuning. If the performance ranking is similar between both
methodologies, this suggests it is sufficient to use linear probing for
benchmarking, while being representative of the performance ranking with
end-to-end fine-tuning.

\textbf{Changing the decoder.} In semantic segmentation, the decoder maps the encoded features to semantic
classes for each pixel in the image. The commonly used Mask2Former
decoder~\cite{cheng2022masked} achieves state-of-the-art performance for multiple
image segmentation tasks by performing mask classification instead of pixel-wise
classification, using a transformer decoder with masked cross-attention. To assess
whether this more advanced decoder impacts the performance ranking, the analysis
compares using a simple linear layer to Mask2Former. If the performance ranking is
similar between these decoders, this suggests it is sufficient to use a
linear decoder for benchmarking, while being representative of the performance
ranking with Mask2Former.

\textbf{Scaling the model.} ViTs have been scaled up to billions of parameters~\cite{dehghani2023scaling}.
The smallest size available for the models in Table~\ref{tab:models} is ViT-B,
with approximately 86 million parameters. Additionally, the models have a
ViT-L counterpart, with approximately 304 million parameters. To assess whether
increasing the model size impacts the performance ranking, the analysis compares the
ViT-B variants to the ViT-L counterparts. If the performance ranking is
similar between these model sizes, this suggests it is sufficient to use
ViT-B for benchmarking, while being representative of the performance ranking
with ViT-L.

\textbf{Varying the patch size.} The ViT processes images by dividing them into non-overlapping patches, which
are then linearly embedded to generate the input sequence for the model.
Smaller patch sizes lead to an increase in the number of tokens for the same
image, enhancing the model's capacity to discern fine-grained details, while
requiring more compute. The models are pretrained with a certain number of
tokens per image (see Table~\ref{tab:models}), and it is unclear to what extent
the models are effective when fine-tuned with a different number of tokens. To
assess whether increasing the number of tokens impacts the performance ranking,
the analysis compares the commonly used $16\times16$ patch size to a smaller
$8\times8$ patch size, which quadruples the number of tokens. If the
performance ranking is similar between these patch sizes, this suggests it is
sufficient to use a $16\times16$ patch size for benchmarking, while being
representative of the performance ranking with an $8\times8$ patch size.

\textbf{Changing the downstream dataset.} Current VFMs require supervised fine-tuning to perform the task of semantic
segmentation effectively. The most commonly used dataset for this task is
ADE20K~\cite{zhou2017scene}, which consists of complex scenes, annotations for
150 semantic classes, and an average image size of around $512\times512$ pixels.
Secondly, another commonly used dataset is PASCAL
VOC~\cite{everingham2010pascal}, which consists of object-centric scenes,
annotations for 21 semantic classes, and an average image size of around
$512\times512$ pixels. Lastly, another commonly used dataset is
Cityscapes~\cite{cordts2016cityscapes}, which consists of homogeneous urban
scenes, fine-grained annotations for 19 semantic classes, and a consistent
image size of $2048\times1024$ pixels. Together, these datasets encompass a selection
of varying scene types, classes, and image sizes. To assess how the choice of
downstream dataset impacts the performance ranking, the analysis compares
training and evaluating on ADE20K to training and evaluating on PASCAL VOC as
well as Cityscapes. If the performance ranking is similar between these
datasets, this suggests it is sufficient to use one of the datasets for
benchmarking, as the ranking for this dataset is representative of the other
ones.

\textbf{Introducing a domain shift.} Recent work~\cite{wei2023stronger} has shown that large VFMs, when fine-tuned
for semantic segmentation, exhibit superior robustness to domain shifts,
surpassing specialized domain generalization methods. To assess whether the
performance ranking is consistent across domain shifts, the analysis compares
training on the Cityscapes training set to training on the synthetic GTA
V~\cite{richter2016playing} dataset, by evaluating both on the Cityscapes
validation set. By training on synthetic data while evaluating on real data, a
significant domain shift is introduced. If the performance ranking is similar
with and without a domain shift, this suggests it is sufficient to only perform
in-distribution evaluation on Cityscapes for benchmarking, as the ranking is representative
of training on data from GTA V and evaluating on out-of-distribution data from Cityscapes.

\textbf{Default setup.} The default benchmarking setup for the impact analysis constitutes end-to-end
fine-tuning with a linear decoder of the ViT-B variants with a $16\times16$
patch size on ADE20K

\subsection{Evaluation metrics}
Following common practice, the mean intersection over union (mIoU) is used as
the evaluation metric for semantic segmentation, which measures the overlap
between predicted and ground truth masks. The mIoU is calculated as the average
of the IoU scores for each class, with higher values indicating better
performance.

To assess how changes in settings affect the performance ranking, the Kendall
rank correlation coefficient~\cite{kendall1938new} is used. This coefficient
measures the similarity between two ranking sequences, ranging from -1 for a
reverse ranking to 1 for an identical ranking. It is given by:
\begin{equation}
    \tau = \frac{2}{n(n-1)} \sum_{i<j} \text{sign}(x_i - x_j) \cdot \text{sign}(y_i - y_j)
\end{equation}
where, in this paper, $n$ is the number of models, $x_i$ and $x_j$ are the ranks
of the $i$-th and $j$-th model in one ranking, and $y_i$ and $y_j$ are their
ranks in the other ranking. The choice of this coefficient is motivated by its sensitivity to
changes in rank order, making it effective at identifying how changes
in settings affect the relative performance of models.

\subsection{Implementation details}
\label{sec:implementation}

The patch embeddings are fixed to the same patch size for all models, by
uniform resizing with the FlexiViT~\cite{beyer2023flexivit} method. Likewise,
the positional embeddings are fixed to the same size, by uniform resizing with
bicubic interpolation. For ADE20K and PASCAL VOC, the crop size is fixed to
$512\times512$ pixels, while for Cityscapes and GTA V, it is fixed to
$1024\times1024$ pixels. In accordance with Mask2Former~\cite{cheng2022masked},
training images undergo horizontal flipping, color jittering, resizing between
0.5 and 2.0 times the original size, padding if needed, and random cropping.
For inference, images have the shortest side resized to the fixed crop size,
with forward passes performed using a sliding window over the proportionally
adjusted longer side. Overlaps in the sliding windows are averaged to combine
the logits. The combined logits are resized back to the original image size
with bilinear interpolation. The argmax function is used to convert the logits
to pixel-wise class predictions.

In accordance with Mask2Former, AdamW~\cite{loshchilov2017decoupled} is used as
the optimizer, with a weight decay of 0.05 and a learning rate of 1e-5 for the
encoder and 1e-4 for the decoder, polynomially decayed with a power of 0.9. The
batch size is set to 1 to make training more accessible by reducing the memory
requirements. Through gradient accumulation the effective batch size is
increased to 16 to align with Mask2Former. The number of training steps is set
to 40,000 for ADE20K, and 20,000 for PASCAL VOC, Cityscapes and GTA V.
Experiments are repeated thrice with different random seeds, and the average
performance is reported along with the standard deviation. The code leverages
\texttt{torch.compile} where possible to improve training efficiency. The
default setup requires approximately five hours to train a single model once on
one NVIDIA A100 GPU.

Integrating Mask2Former, designed for multi-scale features from hierarchical
encoders, with the single-scale outputs of a standard ViT, necessitates two
simplifications. Firstly, the pixel decoder, designed to handle multi-scale
features from a hierarchical encoder, is substituted with a straightforward
linear layer that projects patch tokens from the ViT's final layer to the
decoder embedding dimension. Secondly, the multi-scale features and level
embeddings, used for efficiency specifically for hierarchical encoders, are
replaced by the same single-scale features across the decoder layers. The
ablation study in the Mask2Former paper indicates single-scale feature usage
for cross-attention does not compromise performance~\cite{cheng2022masked}.
\section{Results}
\subsection{Impact of settings}
\label{sec:impact}
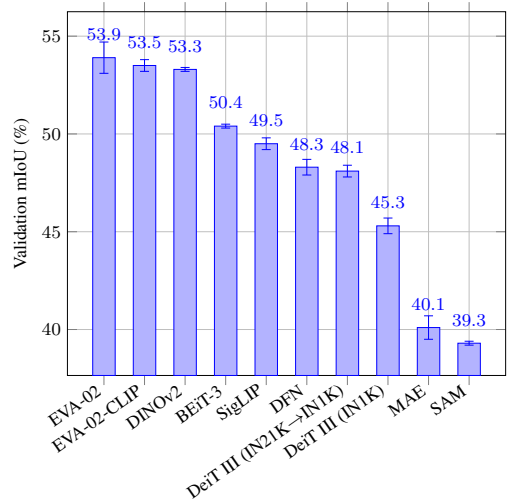
\begin{figure}\footnotesize
    \centering
    \scalebox{0.85}{
        \begin{tikzpicture}
            \begin{axis}[
                    ybar,
                    symbolic x coords={EVA-02, EVA-02-CLIP, DINOv2, BEiT-3, SigLIP, DFN, DeiT III (IN21K$\rightarrow$IN1K), DeiT III (IN1K), MAE, SAM},
                    xtick=data,
                    xticklabel style={rotate=35, anchor=east},
                    nodes near coords,
                    every node near coord/.append style={yshift=5pt},
                    ylabel={Validation mIoU (\%)},
                    grid=major
                ]
                \addplot+[error bars/.cd, y dir=both, y explicit] table [x=Model, y=Performance, y error=StdDev] {\modeltable};
            \end{axis}
        \end{tikzpicture}
    }
    \caption{\textbf{Default setup results.} End-to-end fine-tuning with a linear decoder of the ViT-B variants with a $16\times16$ patch size on ADE20K.}
    \label{fig:default}
\end{figure}

\quad\textbf{Default setup.} The results of end-to-end fine-tuning with a linear decoder of the ViT-B
variants with a $16\times16$ patch size on ADE20K are shown in
Figure~\ref{fig:default}. The subsequent paragraphs detail the impact of
changing individual settings on the performance ranking and training
efficiency, with an overview in Figure~\ref{fig:effect} and
Table~\ref{tab:time}, respectively.

\begin{table}\footnotesize
    \centering
    \begin{tabular}{
            l
            S[table-format=1.1]
            l @{\hspace{3pt}}
            l @{\hspace{3pt}}
        }
        \toprule
        Setting                   & {Training time}       & \multicolumn{2}{c}{Trainable parameters (M)}            \\
        \midrule
        \textcolor{gray}{Default} & \textcolor{gray}{1.0} & \textcolor{gray}{86.6}                       &          \\
        Linear probing            & 0.6                   & 0.1                                          & (-86.5)  \\
        Mask2Former decoder       & 4.1                   & 101                                          & (+14.4)  \\
        ViT-L                     & 1.8                   & 304                                          & (+217.4) \\
        $8\times8$ patch size     & 1.8                   & 88.5                                         & (+1.9)   \\
        \bottomrule
    \end{tabular}
    \caption{\textbf{Training efficiency of benchmark settings.}
        Training time and number of trainable parameters for changed settings compared to the default setup (end-to-end fine-tuning, linear decoder, ViT-B, $16\times16$ patch size).}\label{tab:time}
\end{table}

\pgfplotstableread{
    Model              Performance     StdDev
    EVA-02             53.9            0.8
    EVA-02-CLIP        53.5            0.3
    DINOv2             53.3            0.1
    BEiT-3             50.4            0.1
    SigLIP             49.5            0.3
    DFN                48.3            0.4
        {DeiT III (IN21K$\rightarrow$IN1K)} 48.1            0.3
        {DeiT III (IN1K)}  45.3            0.4
    MAE                40.1            0.6
    SAM                39.3            0.1
}\modeltable

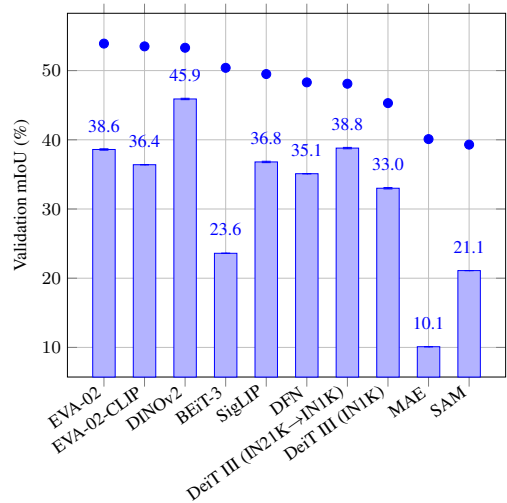
\begin{figure}\footnotesize
    \centering
    \scalebox{0.85}{
        \begin{tikzpicture}
            \begin{axis}[
                    ybar,
                    symbolic x coords={EVA-02, EVA-02-CLIP, DINOv2, BEiT-3, SigLIP, DFN, DeiT III (IN21K$\rightarrow$IN1K), DeiT III (IN1K), MAE, SAM},
                    xtick=data,
                    xticklabel style={rotate=35, anchor=east},
                    nodes near coords,
                    every node near coord/.append style={yshift=5pt},
                    point meta=explicit symbolic,
                    ylabel={Validation mIoU (\%)},
                    grid=major
                ]

                \addplot+[
                    error bars/.cd,
                    y dir=both,
                    y explicit
                ] table [
                        x=Model,
                        y=Performance,
                        y error=StdDev,
                        meta=Performance
                    ] {\modeltable};

                \addplot+[only marks, mark=*, mark options={fill=blue, draw=blue}] table [x=Model, y=Default] {\modeltable};
            \end{axis}
        \end{tikzpicture}
    }
    \caption{\textbf{Linear probing results.} Freezing the encoder results in a correlation coefficient of 0.47 and reduces training time by 0.6 times compared to end-to-end fine-tuning (blue dots).}\label{fig:linear}
\end{figure}

\textbf{Freezing the encoder.} The results comparing linear probing to the default experiment of end-to-end
fine-tuning are shown in Figure~\ref{fig:linear}. All models show
large drops in performance and high variability in the size of
these drops, while training time is reduced by 0.6 times. Current VFMs
may not acquire enough spatial semantic understanding in pretraining to
perform semantic segmentation effectively without end-to-end fine-tuning. Furthermore, a low
correlation coefficient of 0.47 suggests freezing the encoder
significantly changes the performance ranking. Therefore, even though linear
probing is efficient, end-to-end fine-tuning is recommended for a
representative benchmark.

\pgfplotstableread{
    Model           Performance         StdDev      Default
    EVA-02          38.6                0.1         53.9
    EVA-02-CLIP     36.4                0.0         53.5
    DINOv2          45.9                0.1         53.3
    BEiT-3          23.6                0.0         50.4
    SigLIP          36.8                0.1         49.5
    DFN             35.1                0.0         48.3
        {DeiT III (IN21K$\rightarrow$IN1K)}        38.8                0.1         48.1
        {DeiT III (IN1K)}         33.0                0.1         45.3
    MAE             10.1                0.0         40.1
    SAM             21.1                0.0         39.3
}\modeltable

\begin{figure}\footnotesize
    \centering
    \scalebox{0.85}{
        \begin{tikzpicture}
            \begin{axis}[
                    ybar,
                    symbolic x coords={EVA-02, EVA-02-CLIP, DINOv2, BEiT-3, SigLIP, DFN, DeiT III (IN21K$\rightarrow$IN1K), DeiT III (IN1K), MAE, SAM},
                    xtick=data,
                    xticklabel style={rotate=35, anchor=east},
                    nodes near coords,
                    every node near coord/.append style={yshift=5pt},
                    point meta=explicit symbolic,
                    ylabel={Validation mIoU (\%)},
                    grid=major
                ]

                \addplot+[
                    error bars/.cd,
                    y dir=both,
                    y explicit
                ] table [
                        x=Model,
                        y=Performance,
                        y error=StdDev,
                        meta=Performance
                    ] {\modeltable};

                \addplot+[only marks, mark=*, mark options={fill=blue, draw=blue}] table [x=Model, y=Default] {\modeltable};
            \end{axis}
        \end{tikzpicture}
    }
    \caption{\textbf{Mask2Former results.} Using the Mask2Former decoder results in a correlation coefficient of 0.87 and increases training time by 4.1 times compared to a linear decoder (blue dots).}
    \label{fig:decoder}
\end{figure}
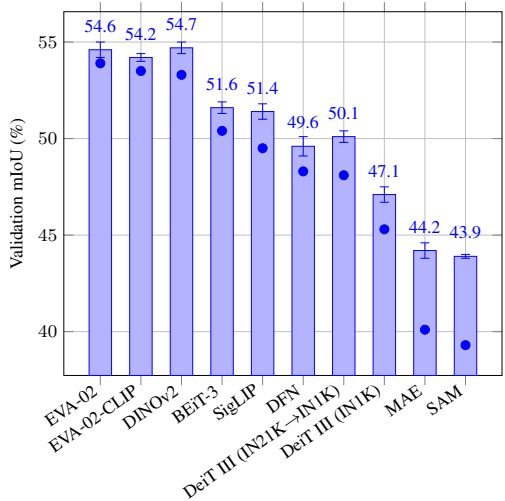

\textbf{Changing the decoder.} The results comparing Mask2Former to the default linear decoder are shown in
Figure~\ref{fig:decoder}. Mask2Former enhances the performance across all
models, at the cost of a training time increase of 4.1 times. A high
correlation coefficient of 0.87 suggests Mask2Former does not significantly
change the performance ranking. The models that did change in ranking did so by
a small margin. Although their ranking remains the same, the lower-performing
models MAE and SAM benefit more, likely because these models require more
adaptation, where Mask2Former provides more capacity for adaptation compared to
a simple linear layer. As the performance ranking remains
largely the same with Mask2Former, while requiring significantly longer training times, the
linear decoder is recommended for an efficient benchmark.

\pgfplotstableread{
    Model           Performance         StdDev      Default
    EVA-02          54.6                0.4         53.9
    EVA-02-CLIP     54.2                0.2         53.5
    DINOv2          54.7                0.3         53.3
    BEiT-3          51.6                0.3         50.4
    SigLIP          51.4                0.4         49.5
    DFN             49.6                0.5         48.3
        {DeiT III (IN21K$\rightarrow$IN1K)}        50.1                0.3         48.1
        {DeiT III (IN1K)}         47.1                0.4         45.3
    MAE             44.2                0.4         40.1
    SAM             43.9                0.1         39.3
}\modeltable

\begin{figure}\footnotesize
    \centering
    \scalebox{0.85}{
        \begin{tikzpicture}
            \begin{axis}[
                    ybar,
                    symbolic x coords={EVA-02, EVA-02-CLIP, DINOv2, BEiT-3, SigLIP, DFN, DeiT III (IN21K$\rightarrow$IN1K), DeiT III (IN1K), MAE, SAM},
                    xtick=data,
                    xticklabel style={rotate=35, anchor=east},
                    nodes near coords,
                    every node near coord/.append style={yshift=5pt},
                    point meta=explicit symbolic,
                    ylabel={Validation mIoU (\%)},
                    grid=major
                ]

                \addplot+[
                    error bars/.cd,
                    y dir=both,
                    y explicit
                ] table [
                        x=Model,
                        y=Performance,
                        y error=StdDev,
                        meta=Performance
                    ] {\modeltable};

                \addplot+[only marks, mark=*, mark options={fill=blue, draw=blue}] table [x=Model, y=Default] {\modeltable};
            \end{axis}
        \end{tikzpicture}
    }
    \caption{\textbf{ViT-L results.} Using the ViT-L counterparts results in a correlation coefficient of 0.87 and increases training time by 1.8 times compared to the ViT-B variants (blue dots).}
    \label{fig:model_size}
\end{figure}
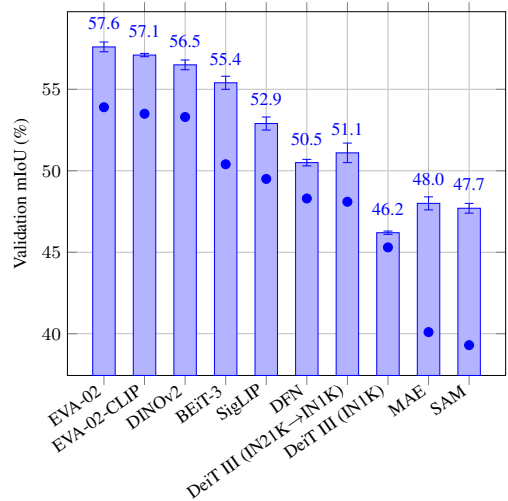

\textbf{Scaling the model.} The results comparing the ViT-L counterparts to the default ViT-B variants of
the models are shown in Figure~\ref{fig:model_size}. Increasing model capacity
enhances the performance across all models, while training time is increased by
1.8 times. A high correlation coefficient of 0.87 suggests increasing model
size does not significantly change the performance ranking. The
lower-performing models MAE and SAM benefit more from a larger model. However,
an outlier is DeiT III (IN1K), which exhibits a minimal performance
improvement. On the other hand, DeiT III (IN21K$\rightarrow$IN1K) benefits
significantly more, implying supervised pretraining on the smaller IN1K dataset
yields limited benefits from scaling the model beyond ViT-B for downstream
semantic segmentation performance. While MAE is pretrained on IN1K, its
improvement may be attributed to the better scaling properties of the
self-supervised MAE objective compared to supervised
learning~\cite{he2022masked}. As the performance ranking remains largely the
same with the ViT-L counterparts, the ViT-B variants of the models are
recommended for an initial benchmark. However, it is important to consider that
scaling the model may reach a plateau in case of limited pretraining data,
depending on the learning objective.

\pgfplotstableread{
    Model           Performance         StdDev      Default
    EVA-02          57.6                0.3         53.9
    EVA-02-CLIP     57.1                0.1         53.5
    DINOv2          56.5                0.3         53.3
    BEiT-3          55.4                0.4         50.4
    SigLIP          52.9                0.4         49.5
    DFN             50.5                0.2         48.3
        {DeiT III (IN21K$\rightarrow$IN1K)}        51.1                0.6         48.1
        {DeiT III (IN1K)}         46.2                0.1         45.3
    MAE             48.0                0.4         40.1
    SAM             47.7                0.3         39.3
}\modeltable

\begin{figure}\footnotesize
    \centering
    \scalebox{0.85}{
        \begin{tikzpicture}
            \begin{axis}[
                    ybar,
                    symbolic x coords={EVA-02, EVA-02-CLIP, DINOv2, BEiT-3, SigLIP, DFN, DeiT III (IN21K$\rightarrow$IN1K), DeiT III (IN1K), MAE, SAM},
                    xtick=data,
                    xticklabel style={rotate=35, anchor=east},
                    nodes near coords,
                    every node near coord/.append style={yshift=5pt},
                    point meta=explicit symbolic,
                    ylabel={Validation mIoU (\%)},
                    grid=major
                ]

                \addplot+[
                    error bars/.cd,
                    y dir=both,
                    y explicit
                ] table [
                        x=Model,
                        y=Performance,
                        y error=StdDev,
                        meta=Performance
                    ] {\modeltable};

                \addplot+[only marks, mark=*, mark options={fill=blue, draw=blue}] table [x=Model, y=Default] {\modeltable};
            \end{axis}
        \end{tikzpicture}
    }
    \caption{\textbf{\boldmath{$8\times8$} patch size results.} Using an $8\times8$ patch size results in a correlation coefficient of 0.78 and increases training time by 1.8 times compared to a $16\times16$ patch size (blue dots).}
    \label{fig:patch_size}
\end{figure}
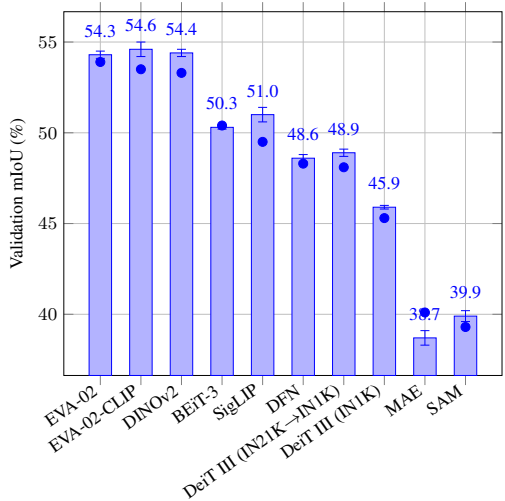

\textbf{Varying the patch size.} The results comparing an $8\times8$ patch size to the default $16\times16$
patch size are shown in Figure~\ref{fig:patch_size}. Most models benefit a
small amount from a smaller patch size, while training time is increased by 1.8
times. A correlation coefficient of 0.78 indicates a small change in the
performance ranking. However, the models that changed in ranking only did so by
a small margin. The number of pretraining tokens per image (see
Table~\ref{tab:models}) might affect how models respond to patch size changes,
although results are inconsistent. Specifically, EVA-02-CLIP shows a greater
improvement with a smaller patch size compared to EVA-02, despite identical
number of pretraining tokens per image. Additionally, an outlier is MAE, being
the only model that shows a decrease in performance with a smaller patch size.
Despite small ranking changes, a smaller patch size does not consistently
advantage all models, with only marginal improvements observed in those that do
benefit. Hence, the most commonly used $16\times16$ patch size is recommended
for an efficient benchmark.

\pgfplotstableread{
    Model           Performance         StdDev      Default
    EVA-02          54.3                0.2         53.9
    EVA-02-CLIP     54.6                0.4         53.5
    DINOv2          54.4                0.2         53.3
    BEiT-3          50.3                0.1         50.4
    SigLIP          51.0                0.4         49.5
    DFN             48.6                0.2         48.3
        {DeiT III (IN21K$\rightarrow$IN1K)}        48.9                0.2         48.1
        {DeiT III (IN1K)}         45.9                0.1         45.3
    MAE             38.7                0.4         40.1
    SAM             39.9                0.3         39.3
}\modeltable

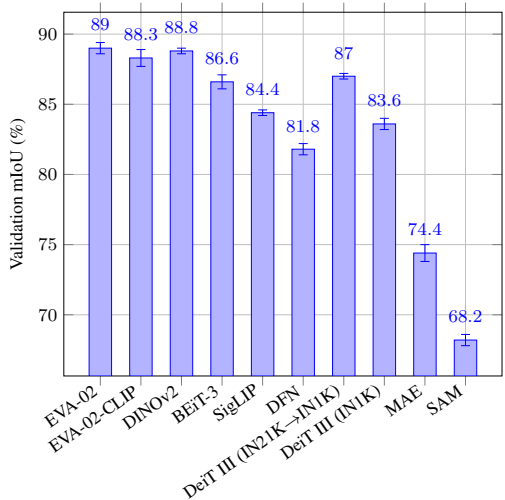
\begin{figure}\footnotesize
    \centering
    \scalebox{0.85}{
        \begin{tikzpicture}
            \begin{axis}[
                    ybar,
                    symbolic x coords={EVA-02, EVA-02-CLIP, DINOv2, BEiT-3, SigLIP, DFN, DeiT III (IN21K$\rightarrow$IN1K), DeiT III (IN1K), MAE, SAM},
                    xtick=data,
                    xticklabel style={rotate=35, anchor=east},
                    nodes near coords,
                    every node near coord/.append style={yshift=5pt},
                    ylabel={Validation mIoU (\%)},
                    grid=major
                ]
                \addplot+[error bars/.cd, y dir=both, y explicit]
                table [x=Model, y=Performance, y error=StdDev] {\modeltable};
            \end{axis}
        \end{tikzpicture}
    }
    \caption{\textbf{PASCAL VOC results.} Using PASCAL VOC results in a correlation coefficient of 0.78 compared to ADE20K.}
    \label{fig:pascal}
\end{figure}

\textbf{Changing the downstream dataset.} The results for PASCAL VOC are shown in Figure~\ref{fig:pascal}. A correlation
coefficient of 0.78 compared to ADE20K indicates a small change in the
performance ranking. The DeiT III models, however, stand out by exhibiting
better relative performance on PASCAL VOC. This enhanced performance is likely
due to their supervised pretraining on the object-centric ImageNet dataset,
which aligns well with the object-centric scenes in PASCAL VOC.

\pgfplotstableread{
    Model           Performance        StdDev
    EVA-02          89.0               0.4
    EVA-02-CLIP     88.3               0.6
    DINOv2          88.8               0.2
    BEiT-3          86.6               0.5
    SigLIP          84.4               0.2
    DFN             81.8               0.4
        {DeiT III (IN21K$\rightarrow$IN1K)}        87.0               0.2
        {DeiT III (IN1K)}         83.6               0.4
    MAE             74.4               0.6
    SAM             68.2               0.4
}\modeltable

\begin{figure}\footnotesize
    \centering
    \scalebox{0.85}{
        \begin{tikzpicture}
            \begin{axis}[
                    ybar,
                    symbolic x coords={EVA-02, EVA-02-CLIP, DINOv2, BEiT-3, SigLIP, DFN, DeiT III (IN21K$\rightarrow$IN1K), DeiT III (IN1K), MAE, SAM},
                    xtick=data,
                    xticklabel style={rotate=35, anchor=east},
                    nodes near coords,
                    every node near coord/.append style={yshift=5pt},
                    ylabel={Validation mIoU (\%)},
                    grid=major
                ]
                \addplot+[error bars/.cd, y dir=both, y explicit]
                table [x=Model, y=Performance, y error=StdDev] {\modeltable};
            \end{axis}
        \end{tikzpicture}
    }
    \caption{\textbf{Cityscapes results.} Using Cityscapes results in a correlation coefficient of 0.56 compared to ADE20K.}
    \label{fig:cityscapes}
\end{figure}
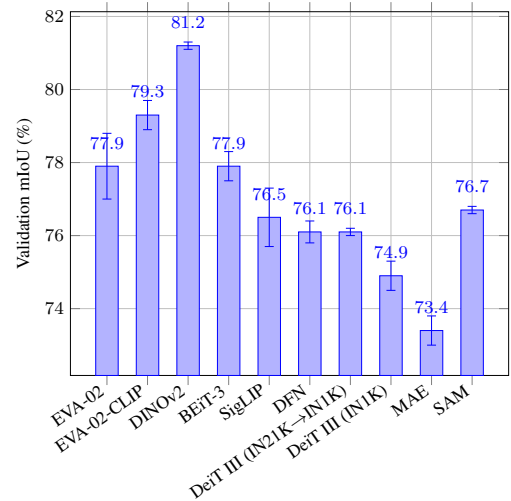

The results for Cityscapes are shown in Figure~\ref{fig:cityscapes}. A
correlation coefficient of 0.56 compared to ADE20K indicates the performance
ranking is even more dissimilar. Notably, DINOv2 and SAM exhibit higher
relative performance on Cityscapes compared to the other datasets. These models
are likely better at capturing the fine-grained details in the Cityscapes
dataset due to their high resolution pretraining in combination with a
patch-level objective, where none of the other models have both of these
properties.

Given the impact of scene similarity and granularity between pretraining and
downstream data on the performance ranking, benchmarking across multiple
datasets is recommended to gain a comprehensive understanding of model
performance in various scenarios.

\pgfplotstableread{
    Model           Performance       StdDev
    EVA-02          77.9              0.9
    EVA-02-CLIP     79.3              0.4
    DINOv2          81.2              0.1
    BEiT-3          77.9              0.4
    SigLIP          76.5              0.8
    DFN             76.1              0.3
        {DeiT III (IN21K$\rightarrow$IN1K)}        76.1              0.1
        {DeiT III (IN1K)}         74.9              0.4
    MAE             73.4              0.4
    SAM             76.7              0.1
}\modeltable

\begin{figure}\footnotesize
    \centering
    \scalebox{0.85}{
        \begin{tikzpicture}
            \begin{axis}[
                    ybar,
                    symbolic x coords={EVA-02, EVA-02-CLIP, DINOv2, BEiT-3, SigLIP, DFN, DeiT III (IN21K$\rightarrow$IN1K), DeiT III (IN1K), MAE, SAM},
                    xtick=data,
                    xticklabel style={rotate=35, anchor=east},
                    nodes near coords,
                    every node near coord/.append style={yshift=5pt},
                    point meta=explicit symbolic,
                    ylabel={Validation mIoU (\%)},
                    grid=major
                ]

                \addplot+[
                    error bars/.cd,
                    y dir=both,
                    y explicit
                ] table [
                        x=Model,
                        y=Performance,
                        y error=StdDev,
                        meta=Performance
                    ] {\modeltable};

                \addplot+[only marks, mark=*, mark options={fill=blue, draw=blue}] table [x=Model, y=Oracle] {\modeltable};
            \end{axis}
        \end{tikzpicture}
    }
    \caption{\textbf{GTA V to Cityscapes results.} Using GTA V for training and Cityscapes for evaluation results in a correlation coefficient of 0.73 compared to using Cityscapes for both training and evaluation (blue dots) and 0.64 compared to ADE20K.}
    \label{fig:gtav}
\end{figure}
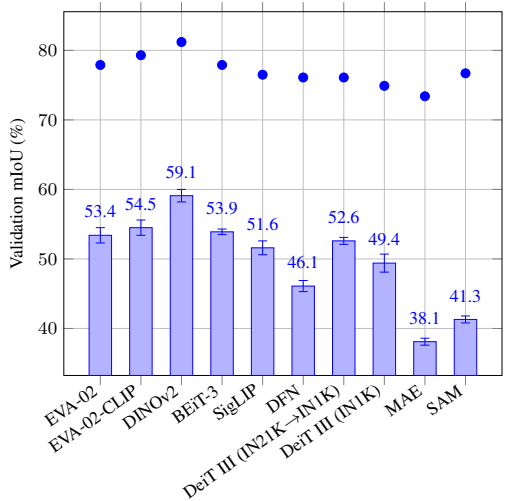

\textbf{Introducing a domain shift.} The results for training on GTA V and evaluating on Cityscapes are shown in
Figure~\ref{fig:gtav}. All models experience a significant decrease in
performance by introducing the synthetic-to-real domain shift. A correlation
coefficient of 0.73 compared to the oracle experiment of using Cityscapes for
both training and evaluation indicates a small change in the performance ranking
when a domain shift is introduced. Notably, SAM and MAE show more substantial
declines, with SAM experiencing the most drastic drop in ranking, falling from
5th to 9th place. This decline may be attributed to the lack of semantic
understanding required for the pretraining tasks of these
models~\cite{wang2023sam}. Given the low semantic complexity of the homogeneous
scenes in the GTA V and Cityscapes datasets~\cite{piva2023empirical}, these
models initially perform well in-distribution. However, when faced with a
domain shift, the lack of semantic understanding becomes apparent. Further
investigation shows similar findings under real-to-real domain shifts.
Therefore, if generalization under domain shifts is important, it is
recommended to include domain shifts in the benchmarking setup, as
in-distribution performance is not necessarily representative of
out-of-distribution performance.

\pgfplotstableread{
    Model           Performance         StdDev      Oracle
    EVA-02          53.4                1.1         77.9
    EVA-02-CLIP     54.5                1.1         79.3
    DINOv2          59.1                0.9         81.2
    BEiT-3          53.9                0.4         77.9
    SigLIP          51.6                1.0         76.5
    DFN             46.1                0.8         76.1
        {DeiT III (IN21K$\rightarrow$IN1K)}        52.6                0.5         76.1
        {DeiT III (IN1K)}         49.4                1.3         74.9
    MAE             38.1                0.5         73.4
    SAM             41.3                0.5         76.7
}\modeltable

\subsection{Analysis of model performance}
\label{sec:analysis}

The benchmarking setup recommended in this paper enables a performance analysis
of VFMs for semantic segmentation.

Surprisingly, SAM, despite being the only model pretrained with mask labels,
performs poorly across most experiments. SAM is pretrained with a promptable
segmentation objective and initialized with the parameters of MAE.
Intriguingly, MAE outperforms SAM in some settings. Further investigation leads
to the observation that, while mask labels arbitrarily belonging to, \eg, whole
objects, parts or subparts, are beneficial for promptable segmentation, their
semantic inconsistency may hinder the learning of effective features for
semantic segmentation. These findings indicate that fine-tuning a model
pretrained for promptable segmentation has limited benefits for semantic
segmentation, and can even lead to negative transfer to this task.

The highest performing models across all experiments performed in this
paper--EVA-02, EVA-02-CLIP, DINOv2 and BEiT-3--have the pretraining objective
of masked image modeling (MIM) with abstract representations in common. While
EVA-02-CLIP directly and EVA-02 and BEiT-3 indirectly rely on weak supervision
from text labels, DINOv2 is pretrained without any labels. EVA-02-CLIP is
pretrained with a language-image learning objective and initialized with the
parameters of EVA-02. Although EVA-02-CLIP did not benefit from additional
language-image learning in the default setup, it outperforms all other
language-image models across all experiments performed in this paper, while it
performs significantly worse than SigLIP and DFN in zero-shot classification on
IN1K (see Table~\ref{tab:reported}). Thus, this shows that zero-shot evaluation
of CLIP models is not indicative of their semantic segmentation performance
downstream, highlighting the necessity of the established benchmark. Moreover,
these findings suggest that MIM with abstract representations is a crucial
pretraining objective for semantic segmentation, even more important than the
type of supervision used.
\section{Conclusion}
This paper presents a study on how to benchmark VFMs for semantic segmentation.
The paper includes an analysis on how varying benchmark settings impact the
performance ranking and training efficiency of VFMs for this task. The results
lead to a recommended benchmarking setup of fine-tuning the ViT-B variants of
VFMs with a $16\times16$ patch size and a linear decoder. Leveraging multiple
datasets for training and evaluation is also recommended, as the performance
ranking across datasets and domain shifts varies. Linear probing is not
recommended, as it is not representative of end-to-end fine-tuning. Finally,
the recommended benchmarking setup enables a performance analysis of VFMs for
semantic segmentation, challenging the value of promptable segmentation
pretraining and highlighting the crucial role of MIM with abstract
representations.
\section{Discussion}
The analysis on how varying benchmark settings impact the performance ranking
of VFMs for semantic segmentation leads to several novel observations on the
performance of specific models in various settings. This facilitates future
work to further investigate the causes of these observations. In addition to
its contributions, this paper has some specific limitations. Firstly, the
hyperparameters used in the experiments mostly align with Mask2Former and the
training steps were determined such that the best performing models converge,
but hyperparameter tuning was not performed. Secondly, although individual
modifications to benchmark settings were investigated, interactions between
settings were not considered, as this would have resulted in a combinatorial
explosion of experiments. Thirdly, the analysis was limited to specific
settings, and the results may not generalize to other settings, such as even
larger models, more limited downstream data, other fine-tuning strategies, or
other types of downstream supervision. Likewise, the impact of different
settings observed in this paper may change for future models, as the field of
VFMs is rapidly evolving. By demonstrating how to evaluate the
representativeness of a benchmarking setup, this paper facilitates future work
to re-evaluate the benchmarking setup for the next generation of VFMs.

{\small
    \paragraph{Acknowledgements.} This paper was supported by Key Digital Technologies
    Joint Undertaking (KDT JU) in EdgeAI “Edge AI Technologies for Optimised
    Performance Embedded Processing” project, grant agreement No 101097300. This
    paper made use of the Dutch national e-infrastructure with the support of the
    SURF Cooperative using grant no.~EINF-5838, which is financed by the Dutch
    Research Council (NWO).}

{
    \small
    \bibliographystyle{ieeenat_fullname}
    \bibliography{main}
}

\end{document}